\documentclass[letterpaper, 10 pt, conference]{ieeeconf}  

\IEEEoverridecommandlockouts                              

\usepackage{graphics} 
\usepackage{epsfig} 
\usepackage{times} 
\usepackage{amsmath} 
\usepackage{amssymb}  
\usepackage{cite}
\usepackage{booktabs} 
\usepackage{makecell}

\usepackage{amsmath,amssymb,amsfonts}
\usepackage{graphicx}
\usepackage{textcomp}
\usepackage{xcolor}
\usepackage{microtype}
\usepackage{booktabs} 
\usepackage{algorithm}
\usepackage{algorithmicx}
\usepackage{algpseudocode}
\usepackage{paralist}
\usepackage{hyperref}
\usepackage{cuted}
\usepackage{float}

\usepackage{colortbl}  
\usepackage{xcolor}

\definecolor{red}{RGB}{170,14,10}
\definecolor{green}{RGB}{83,150,87}
\definecolor{blue}{RGB}{64,224,208}

\title{\LARGE \bf
Cog-GA: A Large Language Models-based Generative Agent for Vision-Language Navigation in Continuous Environments
}

\author{Zhiyuan Li$^{1,2}$, Yanfeng Lu$^{1,2}$,  Yao Mu$^{3}$ and Hong Qiao$^{1,2}$
\thanks{{$^{1}$State Key Laboratory of Multimodal Artificial Intelligence Systems, Institute of Automation, Chinese Academy of Science (CASIA), Beijing 100190, China 
$^{2}$University of Chinese Academy of Sciences (UCAS), Beijing 100049, China
$^{3}$Department of Computer Science, The University of Hong Kong, Hong Kong 999077, Hong Kong Special Administrative Region (SAR)} Correspondence to: Yanfeng Lu {\tt\small<yanfeng.lv@ia.ac.cn>}} 
}

\begin{document}

\maketitle
\thispagestyle{empty}
\pagestyle{empty}

\begin{abstract}

Vision Language Navigation in Continuous Environments (VLN-CE) represents a frontier in embodied AI, demanding agents to navigate freely in unbounded 3D spaces solely guided by natural language instructions. This task introduces distinct challenges in multimodal comprehension, spatial reasoning, and decision-making. To address these challenges, we introduce Cog-GA, a generative agent founded on large language models (LLMs) tailored for VLN-CE tasks.
Cog-GA employs a dual-pronged strategy to emulate human-like cognitive processes. Firstly, it constructs a cognitive map, integrating temporal, spatial, and semantic elements, thereby facilitating the development of spatial memory within LLMs. Secondly, Cog-GA employs a predictive mechanism for waypoints, strategically optimizing the exploration trajectory to maximize navigational efficiency.
Each waypoint is accompanied by a dual-channel scene description, categorizing environmental cues into 'what' and 'where' streams as the brain. This segregation enhances the agent's attentional focus, enabling it to discern pertinent spatial information for navigation. A reflective mechanism complements these strategies by capturing feedback from prior navigation experiences, facilitating continual learning and adaptive replanning.
Extensive evaluations conducted on VLN-CE benchmarks validate Cog-GA's state-of-the-art performance and ability to simulate human-like navigation behaviors. This research significantly contributes to the development of strategic and interpretable VLN-CE agents.

\end{abstract}

\section{Instruction}
Vision Language Navigation (VLN) plays a pivotal role in robotics, where an embodied agent carries out natural language instructions inside real 3D environments based on visual observations. Traditionally, the movements of agents in VLN environments are processed by a pre-prepared navigation graph that the agent traverses. Recognizing this, Krantz et al.\cite{krantz2020beyond} introduced an alternative approach known as Vision-Language Navigation in Continuous Environments (VLN-CE). Unlike traditional methods, VLN-CE eliminates the need for navigation graphs, enabling agents to move freely in 3D spaces. This framework has gained prominence for its realistic and adaptable approach to robotic navigation, allowing agents to respond effectively to verbal commands. Previous works such as ERG\cite{wang2023graph}, VLN-Bridge\cite{hong2022bridging}, and CKR model\cite{gao2021room} primarily focus on reinforcement learning methods. However, reinforcement learning requires lots of interactive data.
\begin{figure}[ht]
\vspace{-0.3cm}
\centerline{\includegraphics[width=0.5\textwidth]{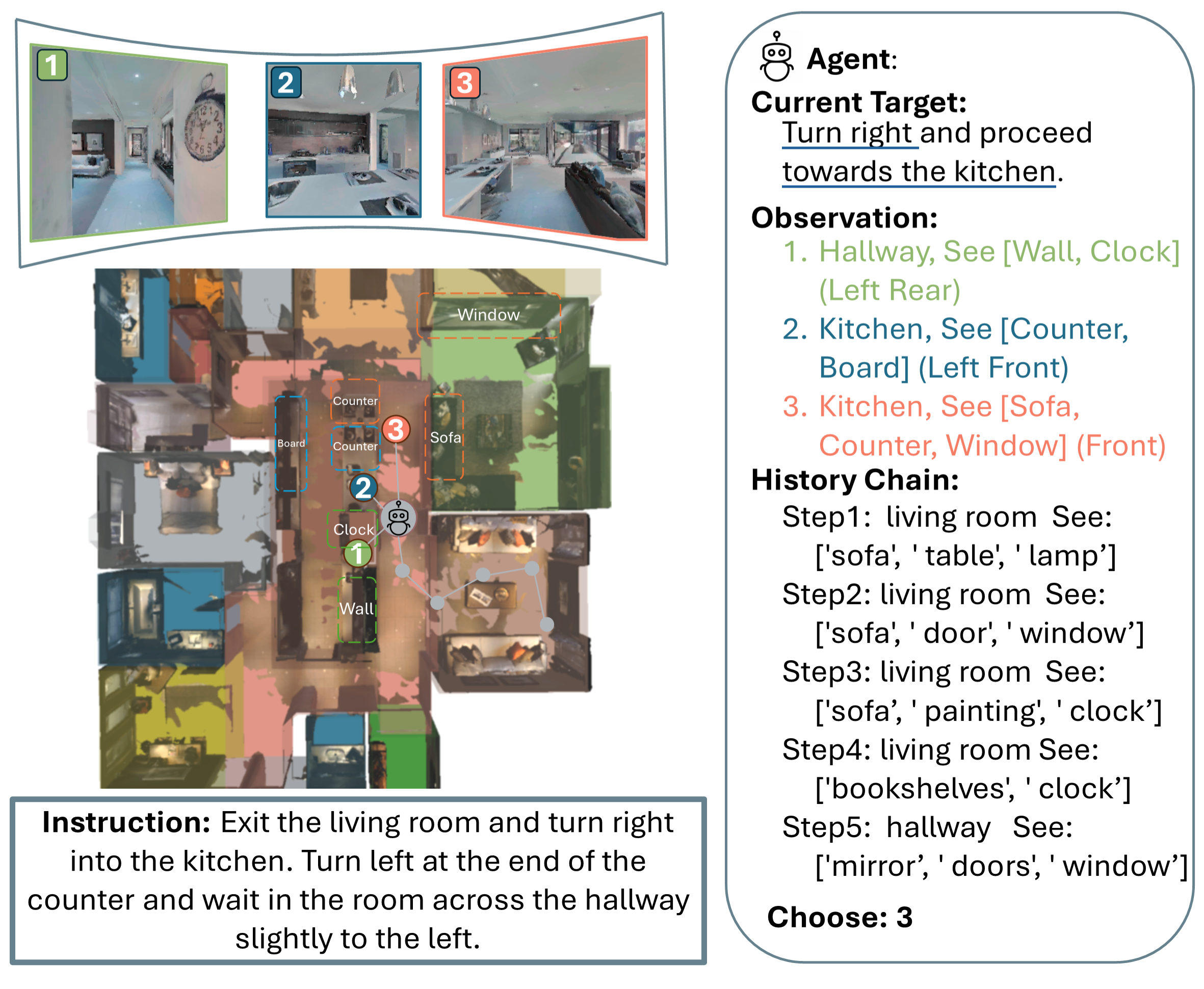}}
\caption{In continuous environments, the Cog-GA agent conducts navigation based on the history chain retrieved from the cognitive map constructed during the navigation process, integrating observations to infer the next target index.}
\label{Fig.1}
\vspace{-0.3cm}
\end{figure}

Large language models (LLMs) have recently illustrated remarkable performance in various fields. Several recent studies have explored the versatility of LLMs in interpreting and navigating complex digital environments, demonstrating their remarkable performance in various fields. For instance, Velma\cite{schumann2023velma} adopts the LLM in Street View VLN tasks. Esc\cite{zhou2023esc} and LFG\cite{shah2023navigation} focus on zero-shot object navigation(ZSON) tasks. ProbES\cite{liang2022visual} further enhances the generalization of LLMs in REVERIE tasks. We aim to leverage the wealth of prior knowledge stored in LLMs to construct an agent with better generalization abilities for VLN-CE tasks. This agent receives dual input from visual and language modalities. It summarizes the key information from the two modalities through its abstract knowledge structures powered by LLMs, bridging sensory modalities and establishing abstract concepts and knowledge structures. 

To this end, we propose Cog-GA (Cognitive-Generative Agent), a LLM-based generative agent for 
vision-language navigation in continuous environments. One of the key challenges in building an efficient VLN agent with LLMs is their lack of inherent spatial memory abilities, as LLMs are trained on flattened text input, lacking the ability to model 3D spatial environments natively. To address this, we introduce the cognitive map, which maintains spatial information related to scene descriptions and landmark objects at each navigation step as a graph. These recorded spatial memories are then retrieved and utilized in subsequent navigation steps. Another core challenge is that valuable waypoints for decision-making by LLMs are often sparsely distributed in the environment. To construct a more reasonable and efficient search space for the agent, we employ the waypoints predictor\cite{hong2022bridging}. For each waypoint, we adopt the dual-channel theory\cite{preston2013interplay, komorowski2013ventral} to describe the observed scene efficiently, which divides scene descriptions into the "what" stream related to landmark objects and the "where" stream concerning spatial characteristics of indoor environments. This division aligns well with the navigation task of reaching objects in different environmental contexts. Since the instructions received by the agent can be separated into sub-instructions corresponding to reaching objects and switching environments, the LLM can effectively focus on the current target. We further introduce a reflection mechanism with a waypoint instruction method to enable the agent to abstract new knowledge from interactions with the environment. 
The LLM then combines these past experiences with the spatial information from the cognitive map to perform more informed navigation planning and facilitate continuous learning and adaptation. Cog-GA employs the LLM to fuse perception results and historical information by maintaining temporal, spatial, and descriptive memories in a cognitive map. 
Each navigation step optimizes the search space using predicted waypoints, abstracting scene descriptions through dual "what" and "where" channels to emphasize relevant objects and spatial contexts. The system learns from experience and adapts its policy, with a reflection mechanism capturing navigation feedback via the LLM.
Extensive experiments on the VLN-CE dataset confirm that our Cog-GA agent achieves promising performance with psychologically human-like behavioral simulation. This work lays a foundation for developing more intelligent, human-like vision-language navigation agents that can strategically adapt to new environments while leveraging prior knowledge from language models. Our key contributions can be summarized as follows:

\begin{compactitem} %
    \item We propose the Cog-GA framework, a generative agent based on large language models (LLMs) for vision-language navigation in continuous environments (VLN-CE), simulating human-like cognitive processes, including cognitive map construction, memory retrieval, and navigation reflection. Experiments demonstrate that Cog-GA achieves a 48\% success rate comparable to the state-of-the-art on the VLN-CE dataset.

    \item We introduce a cognitive map-based memory stream mechanism that stores spatial, temporal, and semantic information, providing contextual knowledge to the LLM to facilitate navigation planning and decision-making.

    \item We introduce a waypoints predictor and a dual-channel ("what" and "where") scene description approach that optimizes the search space, enabling the LLM to focus on current goals. This method significantly improves the navigation success rate.
\end{compactitem} 

\section{Related work}
\label{Related Work}



\subsection{VLN in Continuous Environments}

Visual language navigation (VLN) has gained prominence across natural language processing, computer vision, and robotics. Introduced by Anderson et al. in 2018 with the Room-to-Room (R2R) dataset \cite{anderson2018vision}, VLN has since expanded to include tasks like Touchdown \cite{chen2019touchdown} and REVERIE \cite{qi2020reverie} in diverse environments. Among the numerous methods developed, the Reinforced Cross-Modal Matching (RCM) method has notably surpassed baselines by 10\% in the Success Rate weighted by Path Length metric \cite{Wang2018ReinforcedCM}. The History Aware Multimodal Transformer (HAMT) further advances long-term navigation by effectively integrating historical context \cite{Chen2021HistoryAM}.

In traditional Vision-and-Language Navigation (VLN) tasks, agents are confined to a restricted graph, limiting their applicability to real-world scenarios. Krantz et al.~\cite{krantz2020beyond} expanded the VLN paradigm to continuous environments, enabling agents to navigate through 3D spaces without pre-defined paths, as introduced in the VLN-CE framework. This setup more closely resembles real-world conditions but introduces new challenges. Zhang et al.~\cite{zhang2020diagnosing} noted that visual appearance significantly affects agent performance, highlighting the necessity for models that generalize well in diverse settings. Wang et al.~\cite{wang2020vision} developed the Reinforced Cross-Modal Matching (RCM) approach, which improved navigation performance. Additionally, Guhur et al.~\cite{guhur2021airbert} showed that pretraining on the BnB1 dataset enhances model generalization to new environments.

The \cite{irshad2021hierarchical} model employs a dual-level decision-making process that uses high-level reasoning to match navigation instructions with visual cues and low-level policies for direct agent control. \cite{irshad2022sasra} introduced the Semantically-aware Spatio-temporal Reasoning Agent (SASRA), combining semantic mapping with a hybrid transformer-recurrence architecture to develop a temporal semantic memory, improving spatial and temporal reasoning capabilities in VLN.

To further address the challenges of the VLN-CE task, Hong et al.~\cite{hong2022bridging} demonstrated that navigation via predicted waypoints markedly enhances performance by bridging the discrete-continuous divide. Wang et al.~\cite{wang2023graph} proposed the Environment Representation Graph (ERG), which fortifies the connection between linguistic and environmental data, boosting VLN-CE performance. Chen et al.~\cite{chen2022boosting} introduced the Direction-guided Navigator Agent (DNA), integrating directional cues into the encoder-decoder framework. However, they noted that extensive training is necessary to acquire prior knowledge.

\subsection{Large Language Models Guided Navigation}


Large language models (LLMs) enhance navigation tasks with robust information processing and extensive knowledge. The VELMA model \cite{schumann2023velma} uses LLMs for navigation based on landmarks described in instructions. The Esc model \cite{zhou2023esc} analyzes object and room-level correlations with targets. The LFG model \cite{shah2023navigation} employs chain-of-thought (CoT) reasoning in LLMs for zero-shot object navigation, minimizing irrelevant travel. Cai et al. \cite{cai2023bridging} extend CoT by clustering panoramic images into nodes for strategic navigation decisions. The Prompt-based Environmental Self-exploration (ProbES) \cite{liang2022visual} advances LLM generalization in VLN and REVERIE tasks, showcasing adaptability. Co-NavGPT \cite{yu2023co} utilizes LLMs for collaborative multi-robot navigation, setting midterm goals based on live map data. Integrating additional cognitive processes could further improve these models.

\section{Material and methods}\label{approach}

We leverage the LLM to stimulate the cognitive process of navigation, including creating the cognitive map, instruction understanding, and the reflection mechanism. By introducing LLM, the VLN agent can obtain tremendous prior knowledge, which enables the agent to process tasks effectively. We construct a graph-based cognitive map as external memory to address the LLM lack of long-term and spatial memory. That allows the LLM-based agent to understand and remember the continuous environment.

\subsection{Generative Agent for VLN-CE Tasks}

We categorize the Vision-and-Language Navigation in Continuous Environments (VLN-CE) task into three phases: generating the search space, high-level target planning, and low-level motion generation. Initially, the search space is constructed by segmenting the continuous environment into waypoints, a crucial preprocessing step that simplifies navigation by reducing it to point selection, thereby enhancing efficiency. For high-level target planning, we employ a planner based on large language models (LLMs), which choose waypoints as targets based on current sub-instructions, utilizing spatial memories from the memory stream's cognitive map. The selected waypoint is then forwarded to the motion generator for action execution.

We introduce a generative agent for VLN-CE tasks that comprises a waypoint predictor, memory stream, instruction processing module, high-level planner, and reflection module. The instruction processing module breaks down tasks into manageable sub-instructions, while the waypoint predictor constructs a search space for waypoints at each step using panoramic observations. A scene describer identifies and categorizes the environment into "what" (landmark objects) and "where" (spatial characteristics) streams, enhancing sub-instruction alignment with the environment. These streams and cognitive and reflection memories from the memory stream guide the high-level planner in forming prompts for the large language model (LLM). The planner uses these prompts to identify the target waypoint index relayed to the low-level actuator. During movement, a reflection generator evaluates the navigation results, providing feedback on each step's impact.


\begin{figure}[ht]
\vspace{-0.3cm}
\centerline{\includegraphics[width=0.5\textwidth]{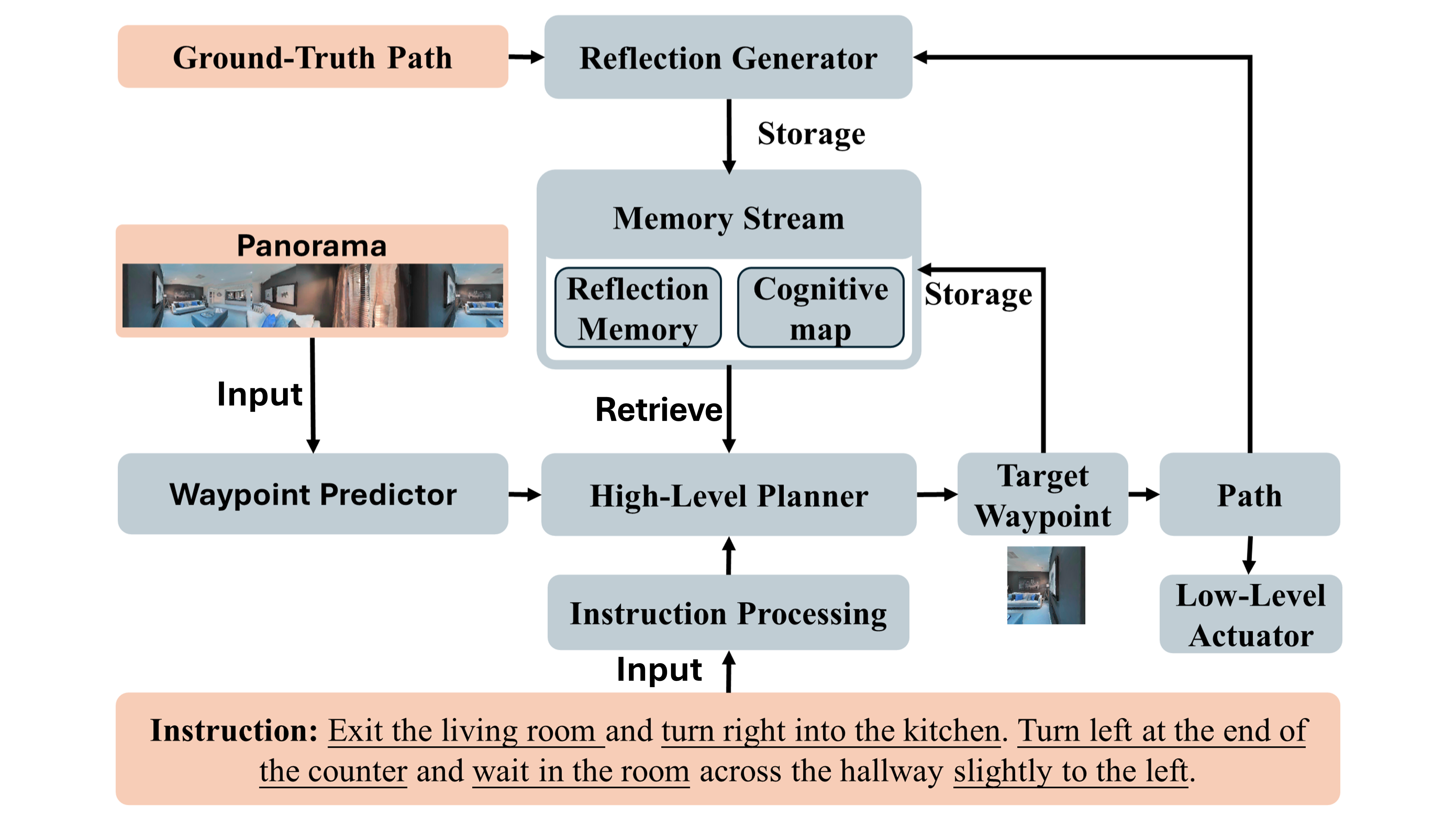}}
\caption{This figure provides an overview of the Cog-GA agent. The panorama of the current position and the main instructions for the task are inputted into both the waypoint predictor module and the instruction processing module. This input is then processed to generate prompts for the LLMs-based high-level target planner, determining the target waypoint. The direction and angle of the target waypoint are subsequently input into the actuator to execute the actions.}
\label{Fig.3}
\vspace{-0.3cm}
\end{figure}

\subsection{Cognitive Map based Target Inference}\label{CM}
Humans and animals create cognitive maps to code, store, and retrieve information about their environments' relative locations and attributes. Introduced by Edward Tolman in 1948 \cite{tolman1948cognitive}, this concept explains how rats learn maze layouts and apply them to humans for navigation and spatial awareness. We use the cognitive map for LLM-based agents and VLN-CE tasks.

A significant challenge for LLMs is the lack of long-term and spatial memory, making external memory crucial \cite{hu2023chatdb, zhu2023ghost}. We address this by introducing a graph-based cognitive map as external memory, which builds and stores spatial memory to help the LLM understand and remember the environment.

The cognitive map starts as an undirected graph $\mathcal{G}(\mathcal{E}, \mathcal{N})$ with nodes $\mathcal{N}_p$ for traversed spaces and $\mathcal{N}_o$ for observed objects. $\mathcal{N}_o$ nodes connect to their corresponding $\mathcal{N}_p$ nodes with 1-weight edges $\mathcal{E}_{p,o}$. Connections between $\mathcal{N}_p$ nodes ($\mathcal{E}_p$) are weighted to represent distance and angle between waypoints, ranging from 0.25 to 3 for distance and 1 to 8 for direction. Each $\mathcal{N}_p$ node also has a time step label $t$. The cognitive map graph is represented as:
\begin{equation}
    \mathcal{G}(\{\mathcal{E}_{p,o}, \mathcal{E}_{p}\}, \{\mathcal{N}_p, \mathcal{N}_o\})
\end{equation}

\begin{figure}[ht]
\centerline{\includegraphics[width=0.45\textwidth]{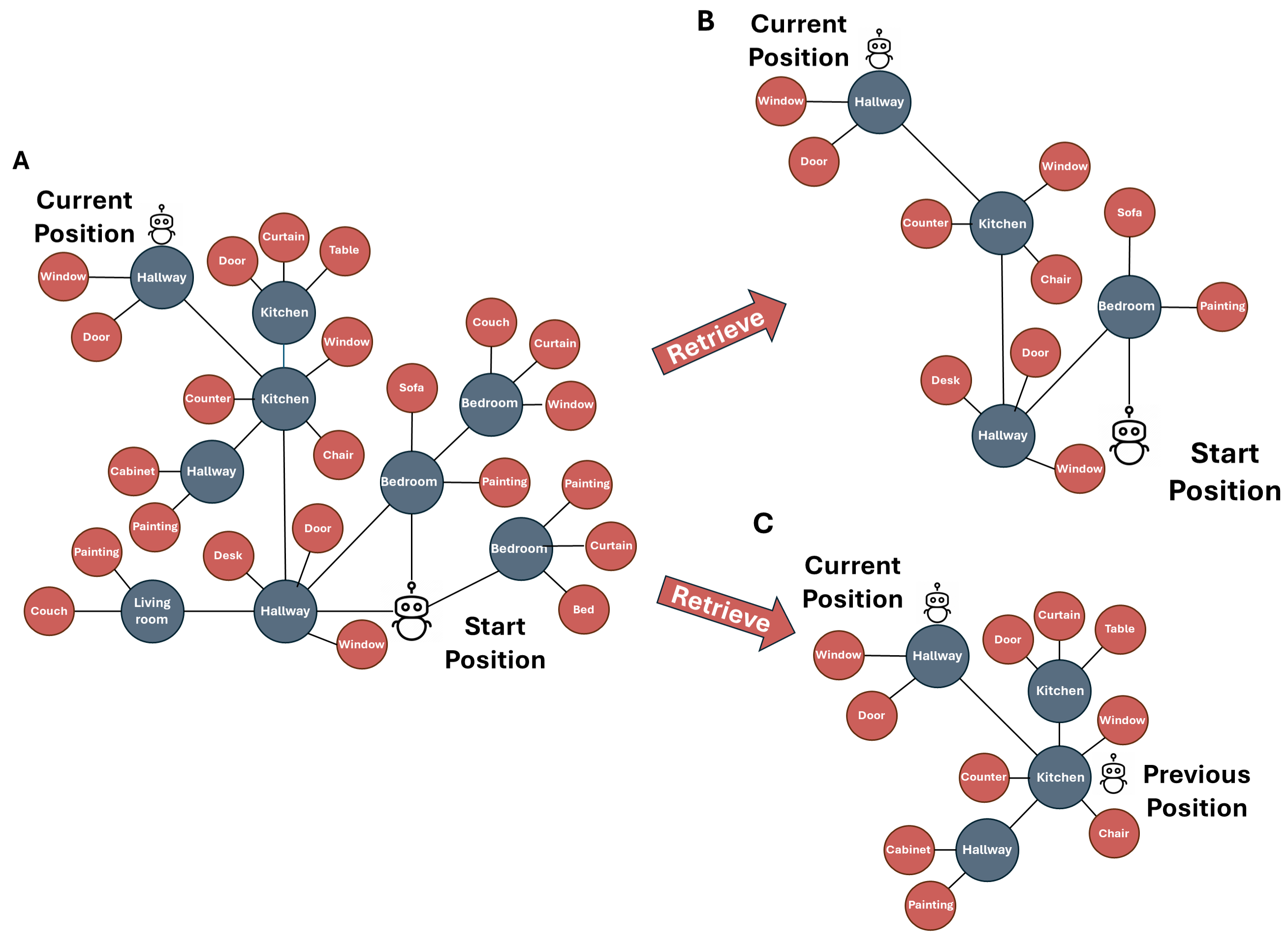}}
\caption{Overview of the agent's cognitive map retrieval methods. Blue nodes are spatially related, and orange nodes are object-related. Connections between blue nodes contain direction and angle data. B represents the history chain, and C the observation chain.}
\label{Fig.4}
\end{figure}

Retrieving the cognitive map is crucial for target inference. As shown in Figure \ref{Fig.4}, we define two retrieval methods: the history and observation chains. The history chain focuses primarily on navigated nodes, providing planners with an abstract view of the current path. In contrast, the observation chain focuses on potential targets between the current and previous positions, offering a broader view of past decisions.

\begin{figure}[ht]
\centerline{\includegraphics[width=0.46\textwidth]{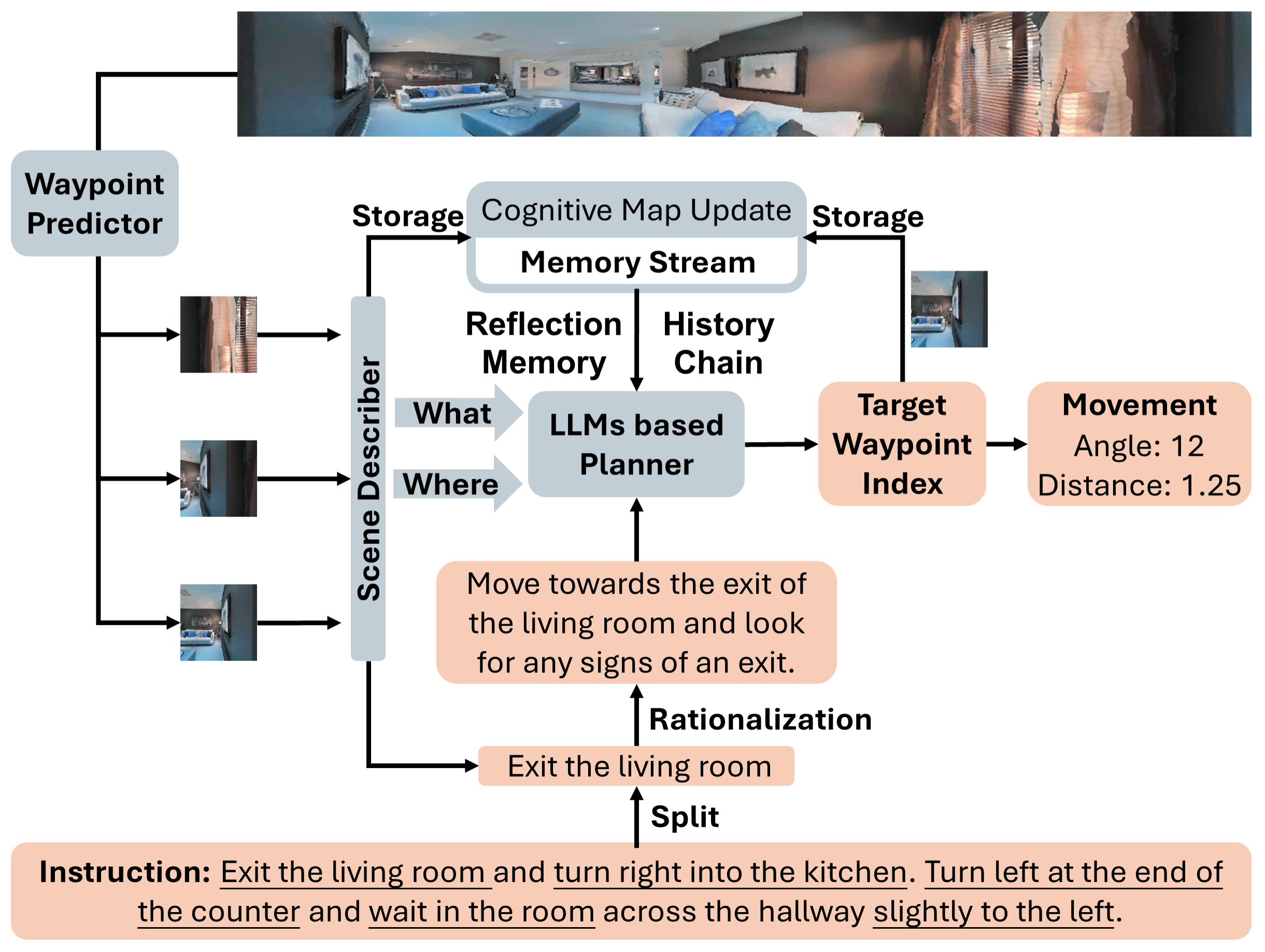}}
\caption{Summary of the agent's inference process. The cognitive map in the memory stream is updated simultaneously.}
\label{Fig.5}
\end{figure}

Figure \ref{Fig.5} outlines the target inference process. After the waypoint predictor segments the panorama, the LLama-based scene describer processes the waypoint image into 'where' and 'what' related words. These waypoints update the cognitive map in the memory stream. The history chain, environment descriptions, reflection memory, and sub-instruction form a unified prompt input to the LLM-based planner. The planner outputs the target waypoint index, which is stored in the memory stream for the cognitive map. The actuator then extracts distance and angle information for the agent's action.

\subsection{Instruction Rationalization based Instruction Processor}\label{IR}
For VLN-CE agents, handling instructions is nontrivial. Using unprocessed instructions directly confuses the planner, causing it to perform meaningless actions. To solve this, we propose an instruction rationalization mechanism. We break the instruction into several sub-instructions using LLMs to guide the agent. However, the original sub-instructions often lack context. For example, the sub-instruction \textit{"Exit the living room."} might confuse the agent about its current target, causing it to repeat routes. Therefore, we include current environment information and the unprocessed instruction to adjust the sub-instruction. 
This process can be expressed as 
\begin{equation}
    I_{i,1} = \mathrm{R}(I_{i,0}|D, \mathcal{I}) \rightarrow ... \rightarrow I_{i,n} = \mathrm{R}(I_{i,n-1}|D, \mathcal{I}) 
\end{equation}
where $I_{i,0}$ is the original sub-instruction, and $\mathcal{I}$ is the unprocessed instruction. 
As the agent moves through the environment, sub-instructions are continuously rationalized. If a sub-instruction is completed, the agent moves to the next one until all sub-instructions are finished. For example, the rationalized sub-instruction \textit{"Find the door of the living room and look for the sign to the kitchen"} is more effective for the agent. At the start of the navigation task, the agent breaks down the natural language instruction into multiple sub-instructions. Each sub-instruction is updated based on observations at each time step as the agent moves, a process we call instruction rationalization.
Detailed discussions of instruction rationalization will be provided in the appendix.

\subsection{Generative Agent with Reflection Mechanism}\label{Rf}

The concept of a generative agent, which blends AI with human-like simulation, represents a significant advancement. These agents mimic human behaviors based on interactions with the environment and past experiences. VLN-CE closely mimics real-world navigation, making it an ideal application for simulating psychological processes during human navigation.

\begin{figure}[ht]
\centerline{\includegraphics[width=0.40\textwidth]{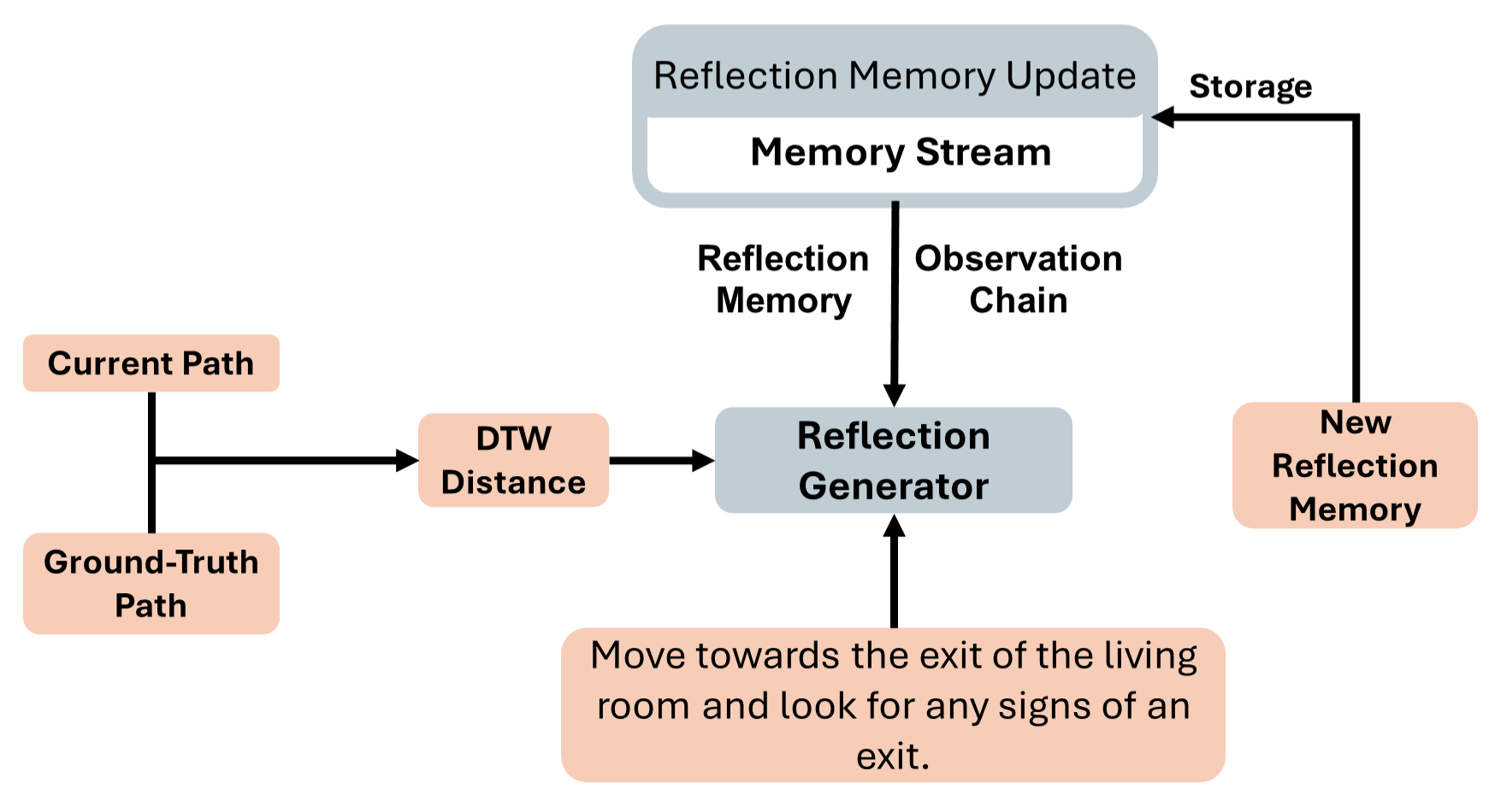}}
\caption{The agent's reflection process is based on ground-truth data. If a new reflection memory is considered non-redundant, it is stored in the memory stream along with its score.}
\label{Fig.6}
\end{figure}

While navigating, the agent receives panoramic waypoint inputs as observations. The scene describer provides a structural description of each waypoint to the planner, along with the history chain from the cognitive map (section \ref{CM}) and reflection memories from the memory stream. The planner uses this information to determine the next navigation waypoint based on the sub-instruction.

After the planner identifies the target, the angle and distance to the waypoint are sent to the low-level actuator to move the agent. The agent then reflects on its movements to gather valuable experiences for future tasks. This reflection helps the agent understand why it succeeded or failed, gaining general knowledge about the environment. However, LLMs can be overwhelmed by too many experiences, disrupting their decision-making. 
We define three parameters for reflection memory in navigation tasks: optimal distance (measured by dynamic time warping between current and correct sequences), proximity (time closeness to the current step), and repeatability (frequency of similar memories). Identical new memories are not stored but update the existing memory's proximity and repeatability.
The score of each reflection memory is defined as follows:
\begin{equation}
    Score_m = \frac{|d_m - \delta|}{\delta} + \frac{t_m}{T} + \frac{r_m}{\max_{r_n \in R}r_n}
\end{equation}
where $d_m$ is the optimal distance, $\delta$ is the threshold parameter of the optimal distance, $t_m$ is proximity, $T$ is the current time step, $r_m$ is repeatability, and $R$ is the set of repeatability of reflection memories. The forgetting process will eliminate reflection memories with scores in the bottom 10\%.


\begin{algorithm}[ht] 
   \caption{Cog-GA}
   \label{alg:example}
\begin{algorithmic}
   \State {\bfseries Input:} Instruction $\mathcal{I}$, Environment $\mathcal{P}$
   \State Initialization: 
   \State \quad Sub-Instruction Set $\mathrm{I}=\{I_{1,0}, I_{2,0},...,I_{n,0}\}$
   \State \quad Cognitive Map $\mathcal{G}(\mathcal{E},\mathcal{N})$, Memory Stream $\mathcal{M}$
   \State \quad $i=1$, $t=1$
   \Repeat
   \State $\mathrm{W} = \{o_1, o_2, ... o_m\}$, $o_k \in waypoint(\mathcal{P})$
   \State $D_k, r_k = Discriber(o_k)$ $k\in{1,2,..m}$
   \State $target = Planner(I_{i,j}, \mathcal{G}, \mathcal{M}, D_{k,k\in{1,2,..m}})$
   \State $y = T(target)$
   \State \textbf{Update} $\mathcal{P}(y)$
   \State \textbf{Update} $\mathcal{G}(\mathcal{E},\mathcal{N})$
   \State $exp = Reflection(y, y^*, \mathcal{G}, \mathcal{M}, I_{i,j})$
   \State \textbf{Update} $\mathcal{M}(exp)$
   \State $I_{i, j+1} = R(I_{i,j}|D_{k,k\in{1,2,..m}},\mathcal{I})$
   \If{$Complete(I_{i,0})$ is $true$}
   \State $i=i+1$
   \EndIf
   \State $t=t+1$
   \If{$Complete(\mathcal{I})$ is $true$}
   \State break
   \EndIf
   \Until{$i=n$}
\end{algorithmic}
\end{algorithm}

\section{Experiments}
To verify the performance of our agent, we deployed our method in VLN-CE environments. This section outlines our experimental setup and implementation details and compares our performance against standard VLN-CE methods. We also highlight several notable features of LLM agents that could inform future research directions. Finally, we assess the impacts of our core methods and provide visual analyses.

\subsection{Experimental Setup}\label{setup}
We conducted experiments on the VLN-CE dataset \cite{krantz2020beyond}, which includes 90 Matterport3D \cite{chang2017matterport3d} scenes. Due to the extended response time of LLaMA, we randomly selected 200 tasks in unseen validation environments for our experiments. Following the methodologies of \cite{krantz2020beyond, krantz2021waypoint}, we used five evaluation metrics \cite{anderson2018evaluation}: Navigation Error (NE), Trajectory Length (TL), Success Rate (SR), Oracle Success Rate (OSR), and Success Rate weighted by Path Length (SPL), with SR being the primary metric.

\subsection{Implementation Details}
We utilize Vicuna-7b \cite{zhu2023minigpt} as the scene describer to align visual modality information with natural language information. For path planning, considering the balance between performance and response time, we adopted GPT-3.5. As used in \cite{hong2022bridging}, the Waypoint Predictor is employed with a candidate waypoint number set to 7. Our experiment is implemented in PyTorch, utilizing the Habitat simulator \cite{savva2019habitat}, LangChain, and trained on two NVIDIA RTX 4090 GPUs.

\subsection{Comparison with Previous VLN-CE Methods}
In line with previous research, we compare our agent with five previously published VLN-CE methods: Waypoint \cite{krantz2021waypoint}, CMA \cite{krantz2020beyond}, BridgingGap \cite{hong2022bridging}, LAW \cite{raychaudhuri2021language}, and Sim2Sim \cite{krantz2022sim}. All experiments were conducted using the same setup. The results are presented in Table \ref{res-table}. Our Cog-GA demonstrated a notable advantage in Success Rate (SR) and Oracle Success Rate (OSR), indicating that the LLM-based agent performs better and effectively transfers its prior knowledge. However, it is essential to note that the trajectory length is significantly higher than other methods. That is attributed to the agent's conservative stopping mechanism, which prefers to get as close to the target point as possible.

\begin{table}[ht]
\caption{Impacts of core method components on the VLN-CE dataset validation (unseen environments, 200 tasks). All setups are configured as described in Section \ref{setup}.}
\label{res-table}
\vskip 0.20in
\begin{center}
\begin{small}
\begin{sc}
\begin{tabular}{lcccccr}
\toprule
Method      & NE $\downarrow$ & TL & SR $\uparrow$ & OSR $\uparrow$ & SPL $\uparrow$ \\
\midrule
Waypoint\cite{krantz2021waypoint}  & 6.31& 7.62& 36& 40& 34 \\
CMA\cite{krantz2020beyond}       & 7.60& 8.27& 29& 36& 27 \\
BridgingGap\cite{hong2022bridging} & 5.74& 12.2& 44& 53& 39 \\
LAW\cite{raychaudhuri2021language}   & 6.83& 8.89& 35& 44& 31 \\
Sim2Sim\cite{krantz2022sim}    & 6.07& 10.7& 43& 52& 36 \\
\rowcolor{lightgray} Cog-GA(Our) & \textbf{5.32}& 18.3& \textbf{48}& \textbf{59}& \textbf{42} \\ 
\bottomrule
\end{tabular}
\end{sc}
\end{small}
\end{center}
\vskip -0.1in
\end{table}


\subsection{Ablation Experiments}
To verify the effectiveness of each component of our method, we conducted ablation experiments based on the validation setup in unseen environments. These experiments focused on the influence of each element on Trajectory Length (TL), Success Rate (SR), and Oracle Success Rate (OSR). Specifically, we examined the reflection mechanism, the instruction rationalization mechanism, and the cognitive map. The results of the ablation experiments are presented in Table \ref{abl-table}.

\begin{table}[ht]
\caption{Ablation experiment results}
\label{abl-table}
\vskip 0.20in
\begin{center}
\begin{small}
\begin{sc}
\begin{tabular}{lcccr}
\toprule
Method          &  SR $\uparrow$ & OSR $\uparrow$ & SPL $\uparrow$ \\
\midrule
(-)reflection      &  41& 57& 38 \\
(-)rationalization &  16& 33& 24 \\
(-)cognitive map   &  22& 46& 32 \\
\rowcolor{lightgray} Cog-GA (Ours)     & \textbf{48}& \textbf{59}& \textbf{42} \\ 
\bottomrule
\end{tabular}
\end{sc}
\end{small}
\end{center}
\vskip -0.1in
\end{table}

The results demonstrate that the instruction rationalization mechanism and the cognitive map significantly influence the agent's performance, while the reflection mechanism has a relatively lower impact. However, all components contribute to the overall effectiveness of the agent. The reflection mechanism, in particular, is primarily used for experience accumulation, suggesting that its importance will grow over the long term as more reflective memory is accumulated.

\section{Conclusion}
In this paper, we introduce a generative agent for VLN-CE that demonstrates the powerful ability of natural language to represent. By mimicking human navigation processes, the agent excels in performance. The simulation of brain navigation could bring an advantage for VLN-CE tasks. The cognitive map-based external memory enables the LLM agent to memorize spatial information. However, communication speed with LLMs is a significant hurdle when using these agents in robotic systems. Future efforts will aim to create a more efficient, high-performing generative agent and improve multimodal large models for vision-language navigation.

\section*{Acknowledgements}
This work is supported by the Strategic Priority Research Program of the Chinese Academy of Sciences under  (Grants XDA0450200, XDA0450202), Beijing Natural Science Foundation (Grant L211023), and National Natural Science Foundation of China (Grants 91948303, 61627808)


\bibliographystyle{IEEEtranS} 
\bibliography{main}







\clearpage

\appendix
\label{Appendix}


\section{Task setup}
\textbf{1. Task Setup} 

In Vision-Language Navigation in Continuous Environments (VLN-CE) \cite{krantz2020beyond}, agents must navigate through unseen 3D environments to specific target positions based on language instructions. These environments are considered as continuous open spaces. The agent selects a low-level action from an action sequence library at each step, given the instruction $\mathcal{I}$ and a $360^\circ$ panoramic RGB-D observation $\mathcal{Y}$. Navigation is successful only if the agent selects a stop within 3 meters of the target location.

Recent VLN-CE solutions \cite{hong2022bridging, krantz2022sim} have adopted a high-level waypoint search space approach. During navigation, the agent utilizes a Waypoint Predictor to generate a heatmap covering 120 angles and 12 distances, highlighting navigable waypoints. Each angle increment is 3 degrees, and the distances range from 0.25 meters to 3.00 meters, with 0.25-meter intervals corresponding to the turning angle and forward step size in the low-level action space. This approach translates the problem of inferring low-level controls into selecting an appropriate waypoint.


\section{Optimal Prompt Mechanism for LLMs in Navigation Tasks}
\textbf{2. Optimal Prompt Mechanism for LLMs in Navigation Tasks}

During the development of the agent, we observed several intriguing features. The structural context is crucial for navigation tasks. To direct the LLM's focus toward navigation-related information, we categorized the information into three distinct types: objects, room types, and directions. Consequently, the context should be structured in the format '\textit{Go (direction), Is (room type), See (objects)}.' Maintaining concise context is essential, as complex and miscellaneous contexts can disrupt the LLM's performance.

For waypoint selection, the clarity of surrounding environment descriptions also plays a significant role. We format the environment descriptions as '\textit{In (direction), See (objects), Is (room type)}.' Clear and concise information reduces unnecessary processing burdens for LLMs and minimizes the risk of irrational outputs due to redundant input. However, a fully structured prompt alone is insufficient for VLN-CE tasks. Original instructions often involve multiple steps, and structural division of sub-instructions can lead to information loss and misdirection. Therefore, we introduced a guidance mechanism to provide structural information for the current target while supplementing sub-instructions. As detailed in section \ref{approach}, we divided sub-targets into 'where' and 'what.' The 'where' targets involve switching environments based on room type, and the 'what' targets involve finding specific objects in the current environment. Thus, we constructed the structural guidance as '\textit{You should try to go (where)}' and '\textit{You should try to find (what)}.' This guidance updates simultaneously with the rationalized sub-instruction to ensure coherence.

\section{The Influence of Instruction Quality and Constructing Better Sub-Instructions}\label{InsDis}
\textbf{3. The Influence of Instruction Quality and Constructing Better Sub-Instructions}

Our experiments highlighted the critical importance of instruction quality on navigation outcomes. This section analyzes how to enhance instruction quality during navigation and explores its implications for future work. As described in section \ref{IR}, splitting instructions into multiple steps and continuously rationalizing each step has proven effective. For example, the rationalized sub-instruction \textit{'Find the living room door and look for a sign to the kitchen.'} yielded better results than the unprocessed sub-instruction \textit{'Exit the living room.'}. This finding reveals an interesting phenomenon: for an agent performing a task, the sequence of separated steps should maintain instructional coherence and constantly adapt the description of the target to the practical environment. Similar phenomena may also occur in human cognitive processes. Furthermore, performance improvements observed before and after splitting the original instruction demonstrate that LLMs have limited capacity to process long-term descriptive instructions. Long-term instructions can easily confuse the LLM by presenting multiple potential targets.

\section{The evaluation metric of reflection memory}
\textbf{4. The evaluation metric of reflection memory}

We define three parameters for each reflection memory: optimal distance, proximity, and repeatability. Optimal distance is the dynamic time warping (DTW) between the current and ground-truth navigation sequences. Repeatability counts how often similar memories occur, and proximity is the time between the memory and the current step. If a new memory is identical to an existing one, it won't be stored, and the current memory's proximity and repeatability will be updated. The score of each reflection memory is defined as follows:
\begin{equation}
    Score_m = \frac{|d_m - \delta|}{\delta} + \frac{t_m}{T} + \frac{r_m}{\max_{r_n \in R}r_n}
\end{equation}
where $d_m$ is the optimal distance, $\delta$ is the threshold parameter of the optimal distance, $t_m$ is proximity, $T$ is the current time step, $r_m$ is repeatability, and $R$ is the set of repeatability of reflection memories. The forgetting process will eliminate reflection memories with scores in the bottom 10\%.

\section{Vision-Language Navigation Task Sample}
\textbf{5. Vision-Language Navigation Task Sample}

For the following figures, the left part is the first view image chosen by the agent, and the right part is the map for the task environment. The 'Action,' 'In,' and 'See' are the linguistic observations and movements for the first view image chosen by the agent.

\textbf{TASK INSTRUCTION:} Exit the living room and turn right into the kitchen. Turn left at the end of the counter and wait in the room across the hallway slightly to the left.

\begin{figure}[H]
\centerline{\includegraphics[width=0.5\textwidth]{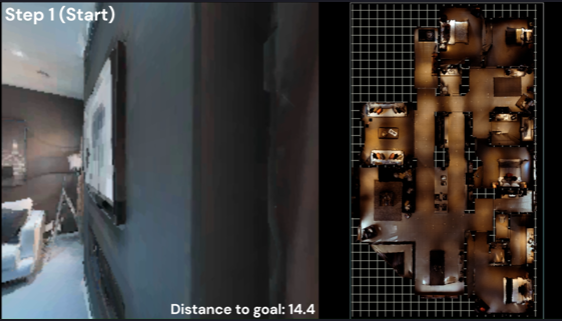}}
\end{figure}

Step: 1 \textcolor{blue}{Action: go Left Front for 0.75 meters} 

\textcolor{red}{In: living room} 

\textcolor{green}{See: ['sofa', ' picture frame', ' lamp', ' bookshelf', ' window', ' clock', ' rug', ' painting', ' curtains', ' table']}

\begin{figure}[H]
\centerline{\includegraphics[width=0.5\textwidth]{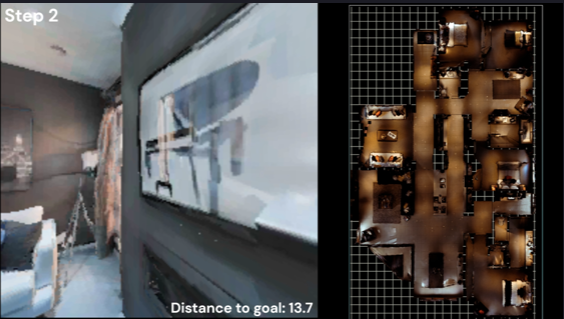}}
\end{figure}

Step: 2 \textcolor{blue}{Action: go Left Front for 1.25 meters} 

\textcolor{red}{In: living room} 

\textcolor{green}{See: ['grey wall', ' white couch', ' table', ' flat screen tv', ' white desk', ' window', ' fireplace', ' carpet', ' black tiles', ' lamp', ' mirror', ' plants']}

\begin{figure}[H]
\centerline{\includegraphics[width=0.5\textwidth]{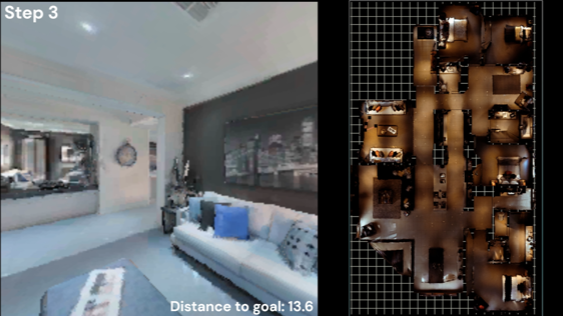}}
\end{figure}

Step: 3 \textcolor{blue}{Action: go Left Front for 2.5 meters} 

\textcolor{red}{In: living room} 

\textcolor{green}{See: ['couch', ' chairs', ' window', ' mirror', ' floor', ' walls', ' table', ' chairs', ' window', ' table']}

\begin{figure}[H]
\centerline{\includegraphics[width=0.5\textwidth]{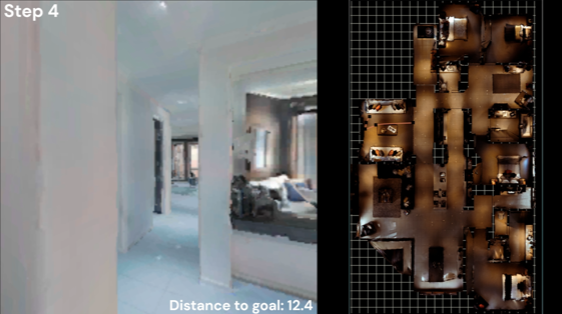}}
\end{figure}

Step: 4 \textcolor{blue}{Action: go Front for 1.75 meters}	 

\textcolor{red}{In: living room} 	 

\textcolor{green}{See: ['mirror', ' furniture', ' living room', ' spacious', ' white tiles', ' large mirror']}

\begin{figure}[H]
\centerline{\includegraphics[width=0.5\textwidth]{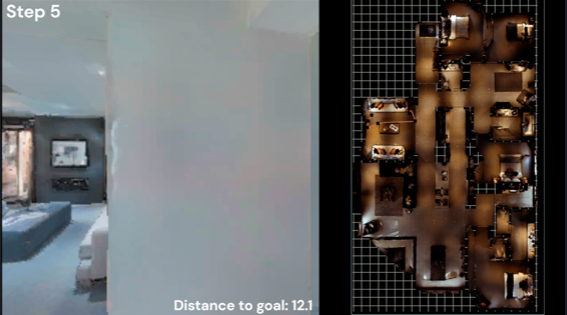}}
\end{figure}

Step: 5 \textcolor{blue}{Action: go Right Rear for 1.75 meters} 

\textcolor{red}{In: living room}

\textcolor{green}{See: ['sofa', ' chandelier', ' fireplace', ' bookshelf', ' painting', ' rug', ' coffee table', ' curtains', ' lamp', ' mirror']}

\begin{figure}[H]
\centerline{\includegraphics[width=0.5\textwidth]{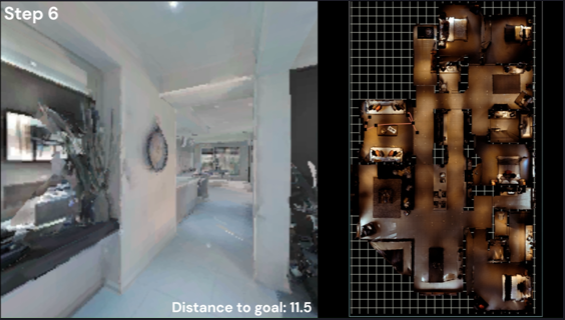}}
\end{figure}

Step: 6 \textcolor{blue}{Action: go Right Front for 1.5 meters}

\textcolor{red}{In: kitchen}

\textcolor{green}{See: ['marble floor', ' blue tiles', ' grey walls', ' white refrigerator', ' silver handle', ' black stove', ' silver microwave', ' silver cabinet doors', ' white handles', ' white dishwasher', ' large mirror', ' silver frame', ' window']}

\begin{figure}[H]
\centerline{\includegraphics[width=0.5\textwidth]{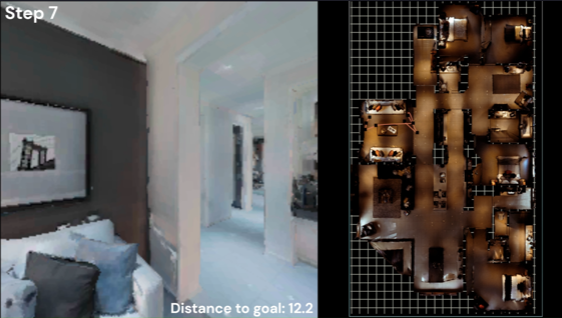}}
\end{figure}

Step: 7 \textcolor{blue}{Action: go Left Rear for 2.0 meters}

\textcolor{red}{In: living room}

\textcolor{green}{See: ['couch', ' coffee table', ' windows', ' wall', ' pictures', ' sink', ' counter', ' cabinets', ' floor', ' rug']}

\begin{figure}[H]
\centerline{\includegraphics[width=0.5\textwidth]{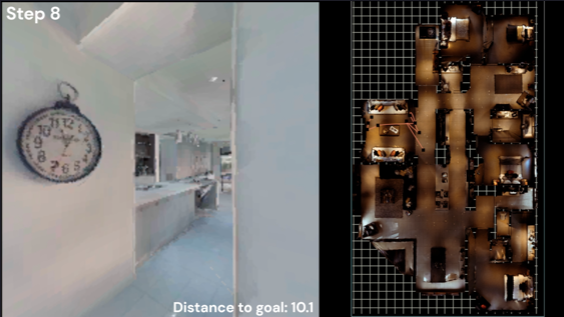}}
\end{figure}

Step: 8 \textcolor{blue}{Action: go Front for 1.25 meters}

\textcolor{red}{In: kitchen}

\textcolor{green}{See: ['clock', ' hallway', ' kitchen', ' tiles', ' floor', ' curtains', ' walls', ' countertop', ' appliances', ' stove']}

\begin{figure}[H]
\centerline{\includegraphics[width=0.5\textwidth]{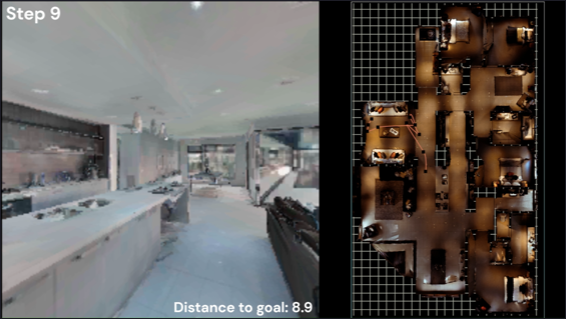}}
\end{figure}

Step: 9 \textcolor{blue}{Action: go Front for 2.0 meters}

\textcolor{red}{In: kitchen} 

\textcolor{green}{See: ['kitchen', ' island counter', ' appliances', ' white walls', ' floor tiles', ' living area']}

\begin{figure}[H]
\centerline{\includegraphics[width=0.5\textwidth]{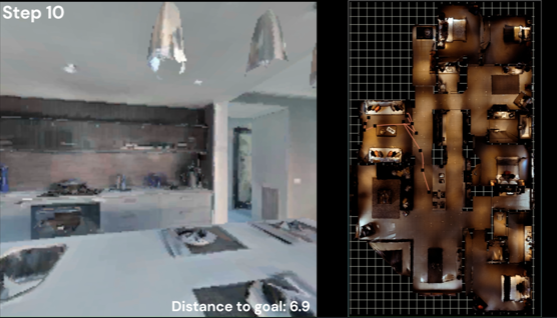}}
\end{figure}

Step: 10 \textcolor{blue}{Action: go Right Front for 1.0 meters}

\textcolor{red}{In: living room}

\textcolor{green}{See: ['sofa', ' door', ' window', ' table', ' chair', ' lamp', ' bookshelf', ' painting', ' rug', ' clock']}

\begin{figure}[H]
\centerline{\includegraphics[width=0.5\textwidth]{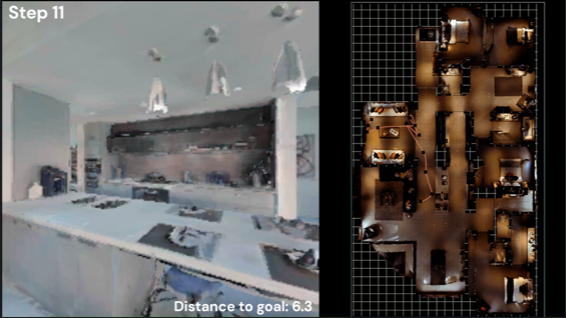}}
\end{figure}

Step: 11 \textcolor{blue}{Action: go Right Rear for 2.0 meters}

\textcolor{red}{In: kitchen}

\textcolor{green}{See: ['kitchen', ' center island, ' counter top', ' stools', ' refrigerator', ' oven', ' dishwasher', ' sink', ' table', ' lamp']}

\begin{figure}[H]
\centerline{\includegraphics[width=0.5\textwidth]{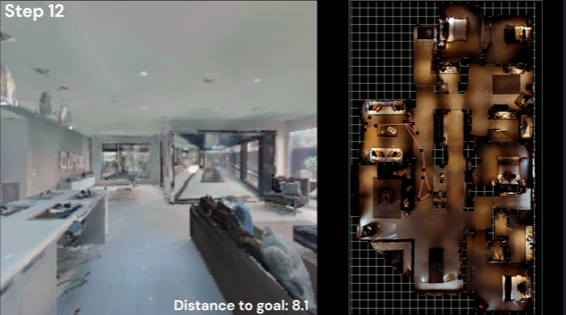}}
\end{figure}

Step: 12 \textcolor{blue}{Action: go Behind for 1.75 meters}

\textcolor{red}{In: living room}

\textcolor{green}{See: ['cabinets', ' countertops', ' windows', ' sofa', ' coffee table', ' dining table']}

\begin{figure}[H]
\centerline{\includegraphics[width=0.5\textwidth]{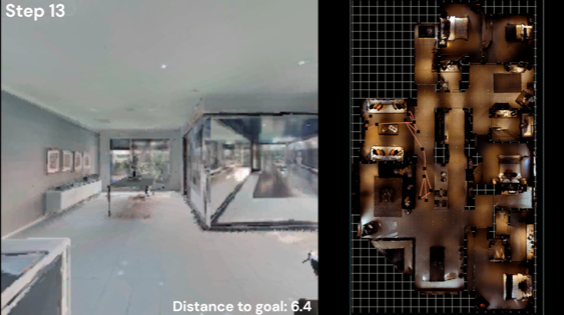}}
\end{figure}

Step: 13 \textcolor{blue}{Action: go Front for 1.5 meters}

\textcolor{red}{In: kitchen}

\textcolor{green}{See: ['windows', ' door', ' counter', ' mirror', ' sink', ' refrigerator', ' chairs', ' coffee table', ' bright', ' spacious']}

\begin{figure}[H]
\centerline{\includegraphics[width=0.5\textwidth]{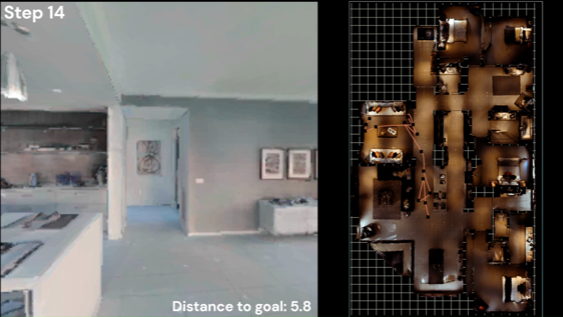}}
\end{figure}

Step: 14 \textcolor{blue}{Action: go Right Side for 1.75 meters}

\textcolor{red}{In: kitchen}

\textcolor{green}{See: ['kitchen island', ' marble countertops', ' induction cooktop', ' chairs', ' paintings', ' walls', ' flooring', ' tile']}

\begin{figure}[H]
\centerline{\includegraphics[width=0.5\textwidth]{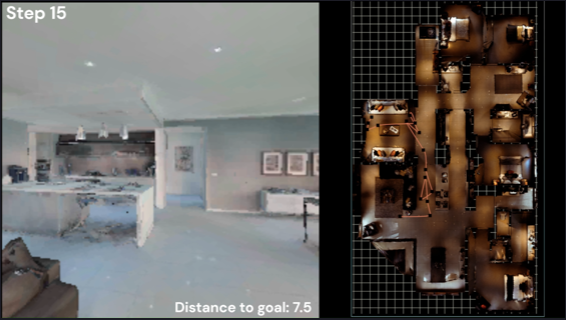}}
\end{figure}

Step: 15 \textcolor{blue}{Action: go Behind for 2.25 meters}

\textcolor{red}{In: living room}

\textcolor{green}{See: ['sofa', ' door', ' window', ' table', ' chair', ' lamp', ' painting', ' bookshelf', ' rug', ' clock']}

\begin{figure}[H]
\centerline{\includegraphics[width=0.5\textwidth]{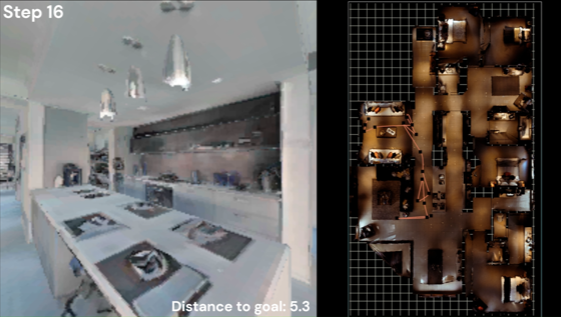}}
\end{figure}

Step: 16 \textcolor{blue}{Action: go Right Front for 2.0 meters}

\textcolor{red}{In: living room}

\textcolor{green}{See: ['sofa', ' door', ' window', ' table', ' chair', ' lamp', ' bookshelf', ' painting', ' rug', ' clock']}

\begin{figure}[H]
\centerline{\includegraphics[width=0.5\textwidth]{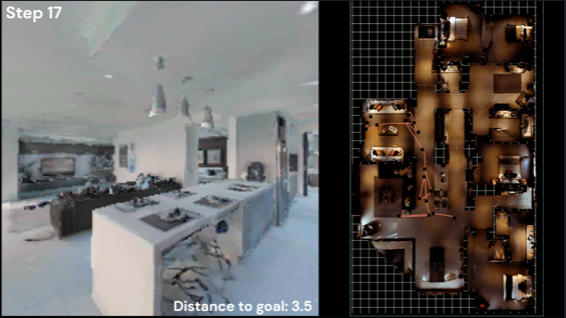}}
\end{figure}

Step: 17 \textcolor{blue}{Action: go Right Rear for 2.0 meters}

\textcolor{red}{In: kitchen}

\textcolor{green}{See: ['kitchen counters', ' sink', ' stools', ' refrigerator', ' oven', ' stove', ' curtains', ' tiles', ' dining table', ' chairs']}

\begin{figure}[H]
\centerline{\includegraphics[width=0.5\textwidth]{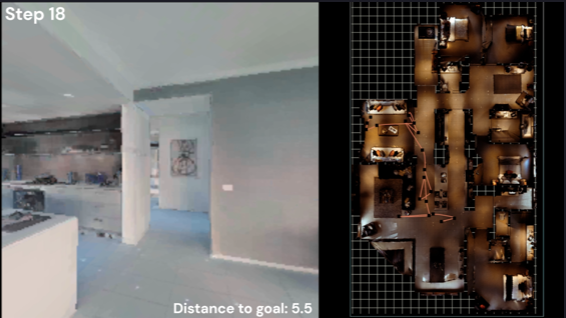}}
\end{figure}

Step: 18 \textcolor{blue}{Action: go Behind for 2.25 meters}

\textcolor{red}{In: kitchen} 

\textcolor{green}{See: ['kitchen', ' counter top', ' subway tiles', ' island', ' stools', ' fridge', ' oven', ' microwave', ' dishwasher', ' cabinets']}

\begin{figure}[H]
\centerline{\includegraphics[width=0.5\textwidth]{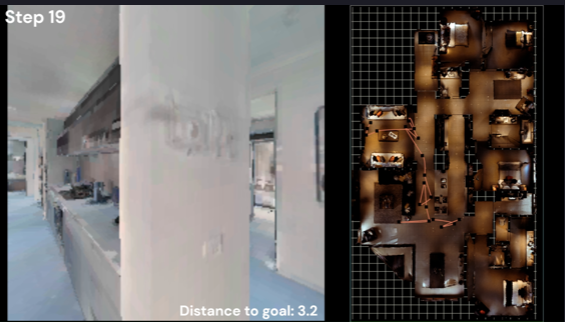}}
\end{figure}

Step: 19 \textcolor{blue}{Action: go Front for 1.25 meters}

\textcolor{red}{In: living room}

\textcolor{green}{See: ['kitchen', ' dining area', ' living area', ' sofa', ' door', ' window', ' table', ' chair', ' stove', ' refrigerator']}

\begin{figure}[H]
\centerline{\includegraphics[width=0.5\textwidth]{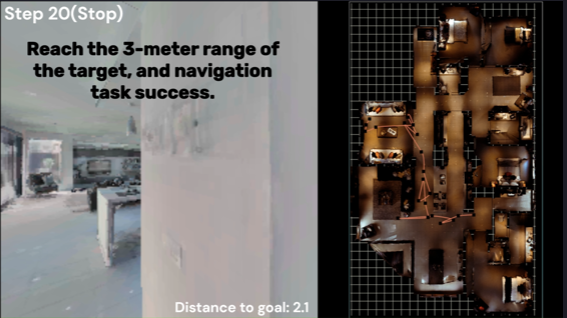}}
\end{figure}

Step: 20 \textcolor{blue}{Action: go Right Rear for 2.25 meters}

\textcolor{red}{In: living room}

\textcolor{green}{See: ['sofa', ' door', ' window', ' table', ' chair', ' lamp', ' bookshelf', ' painting', ' rug', ' clock']}

\begin{figure}[H]
\centerline{\includegraphics[width=0.5\textwidth]{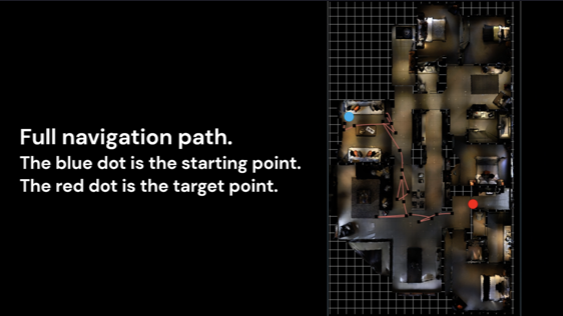}}
\end{figure}

\end{document}